\def\year{2017}
\documentclass[letterpaper]{article}
\usepackage{aaai}
\usepackage{mathpartir}
\usepackage{times}
\usepackage{helvet}
\usepackage{courier}
\usepackage{amsmath}
\usepackage{amssymb}
\usepackage{graphicx}
\usepackage{color}
\usepackage{comment}
\usepackage{lettrine} 
\usepackage[font=small,skip=8pt]{caption}
\usepackage{float}
\makeatletter
\newcommand{\verbatimfont}[1]{\renewcommand{\verbatim@font}{\ttfamily#1}}
\makeatother
\title{Sentiment Transfer using Seq2Seq Adversarial Autoencoders}
\author{Ayush Singh, Ritu Palod\\
\{singh.ay, palod.r\}@husky.neu.edu\\
Northeastern University
}

\begin{document}
%
\maketitle

\section*{Abstract}
Expressing in language is subjective. Everyone has a different style of reading and writing, apparently it all boil downs to the way their mind understands things (in a specific format).
Language style transfer is a way to preserve the meaning of a text and change the way it is expressed. 
Progress in language style transfer is lagged behind other domains, such as computer vision, mainly because of the lack of parallel data, use cases, and reliable evaluation metrics. In response to the challenge of lacking parallel data, we explore learning style transfer from non-parallel data. We propose a model combining seq2seq, autoencoders, and adversarial loss to achieve this goal. The key idea behind the proposed models is to learn separate content representations and style representations using adversarial networks. Considering the problem of evaluating style transfer tasks, we frame the problem as sentiment transfer and evaluation using a sentiment classifier to calculate how many sentiments was the model able to transfer. We report our results on several kinds of models.



\section*{Related Work}
\subsubsection{Style Transfer in Computer Vision} ~\\
Style transfer has made significant progress in computer vision in recent years. \cite{gatys2016image} separated the content and style of images to recombine them generating new images using a linear model to change the color of the pictures. Their methods use only one image to represent a style. However, it does not work for text because a single sentence or short article does not store enough style information.

\cite{zhu2017unpaired} proposes CycleGAN to do image-image translation. It firstly learns a mapping $G : X \rightarrow Y$ using an adversarial loss, and then a reverse mapping $F : Y \rightarrow X$ with a cycle loss $F(G(X)) \approx X$ which performs unpaired image to image translation. CycleGAN shows qualitative results, nevertheless, cycle training is hard to implement with discrete text.

\subsubsection{Style Transfer in Natural Language Processing} ~\\
\cite{DBLP:journals/corr/JhamtaniGHN17} explores automatic methods to transform text from modern English to Shakespearean English using parallel data. The model was based on seq2seq and enriched it with pointer network \cite{vinyals2015pointer}, however, paired word dictionary is a scarce resource that does not exist in most style transfer tasks, and it required parallel corpora.

\cite{fu2017style} proposed a variational auto-encoder (VAE) based model to revise a new sequence to improve its associated outcome. However, there is no significant evaluation for style transfer. It uses nonparallel data. \cite{shen2017style} explored style transfer for sentiment modification, decipherer word substitution ciphers and recovery of word order. They used VAE as the base model and used an adversarial network to align different styles. However, their evaluation only considered the classification accuracy.

There are similar supervised exploratory works by \cite{ficler2017controlling} experiment with controlling several stylistic aspects of the generated text, in addition to its content. The method is based on conditioned recurrent neural networks (CRNN) language model, where the desired content as well as the stylistic parameters serve as conditioning contexts but use hand labeled features and is a supervised task. \cite{li2016persona} encodes personas in distributed embeddings that capture individual characteristics such as background information and speaking style. But again, they use twitter conversations (parallel text) to learn responses. These works are not directly applicable in our case since we want to learn the style and content in an unsupervised manner and using non-parallel text.

\section*{Dataset}
We are using the Yelp dataset.

\begin{enumerate}
\item A large-scale dataset (4.7 million reviews) suitable for effectively training neural networks.
\item Crowd-sourced to collect the natural language reviews written by human beings from over four continents, that avoids over-fitting and improves generalization.
\item  Text reviews are correlated by the stars given by a user to a business location, which is perfect for our case since we can separate styles based on sentiment of the text.
\item To make data suitable for sentiment analysis, we used all reviews with stars more than 3 as \textit{positive}, less than 3 as \textit{negative} and equal to 3 as \textit{neutral}.
\end{enumerate}

\begin{figure}[H]
\begin{minipage}{0.5\textwidth}
\centering
\hspace*{-0.6cm}
\includegraphics[width=\textwidth]{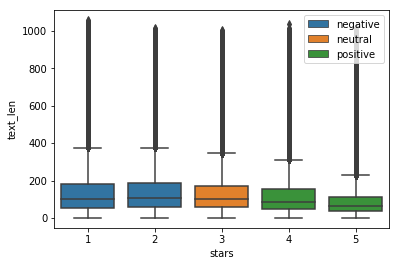}
\caption{Distribution of lengths of different sentiments of reviews}
\end{minipage}
\end{figure}

\subsection*{Preprocessing}
Original dataset came serialized in \textit{json} which we stored in MongoDB and then performed all our exploratory data analysis from database querying and used keras to create our data preprocessing pipeline from loading text to creating vocabulary, encoding into word index sequences and finally padding to fixed length dynamically when creating batches all comes as a prerequisite to feeding data to neural networks.

\begin{figure}[H]
\begin{minipage}{0.5\textwidth}
\centering
\hspace*{-0.6cm}
\includegraphics[width=\textwidth]{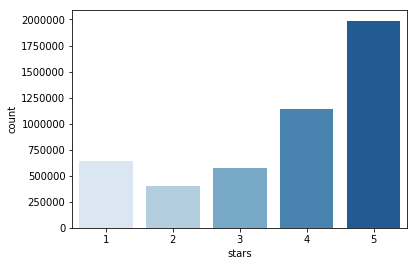}
\caption{Plot of class imbalance in the dataset}
\end{minipage}
\end{figure}

\begin{enumerate}
\item Converted all characters to lower cases.
\item Replaced all the numbers with a special string \_NUM\_.
\item Remove punctuations and special symbols.
\item Remove Neutral reviews (Stars = 3)
\item Maximum Sentence Length 10 words as we are more interested in tips style of reviews e.g. "good Italian food" than longer descriptive ones.
\item Only keep 10k most frequent words in vocabulary and then split the dataset in positive and negative so that both of them share the same vocabulary.
\item The dataset had a class imbalance where positive reviews were 3M vs 1M negative reviews. After preprocessing we were left with 600k reviews, we then handled the class imbalance splitting almost equally.
\item The data is split so that the training (80\%), dev (10\%), and test (10\%).
\end{enumerate}

\section*{Model}
We propose seq2seq autoencoder model and try to train the same model adversarially to separate style from content for style transfer in this paper and evaluate it on various configurations. We try to maximize learning the style of a text and keep the vocabulary consistent among the different styles we want to transfer in. The common ground of the two models is to learn a representation for the input sentence that only contains the content information. Figure 1 illustrates the two models. We give more details about each model in the following sections.

\subsection*{Background: Autoencoders}
\cite{rumelhart1985learning} proposed autoencoders to learn lossy abstract representations of higher dimensional data and were later found to efficient in compressing data. It was mainly used for dimension reduction in the past, but more recently, the concepts have been widely used for generative models. They consist of an encoder that encodes information in a latent space followed by a decoder that tries to regenerate original data using latent space. The learning takes place using gradient descent and backpropagation over a certain number of epochs until network reaches a minimal loss. 
\begin{figure}[H]
\begin{minipage}{0.48\textwidth}
\centering
\includegraphics[width=\textwidth]{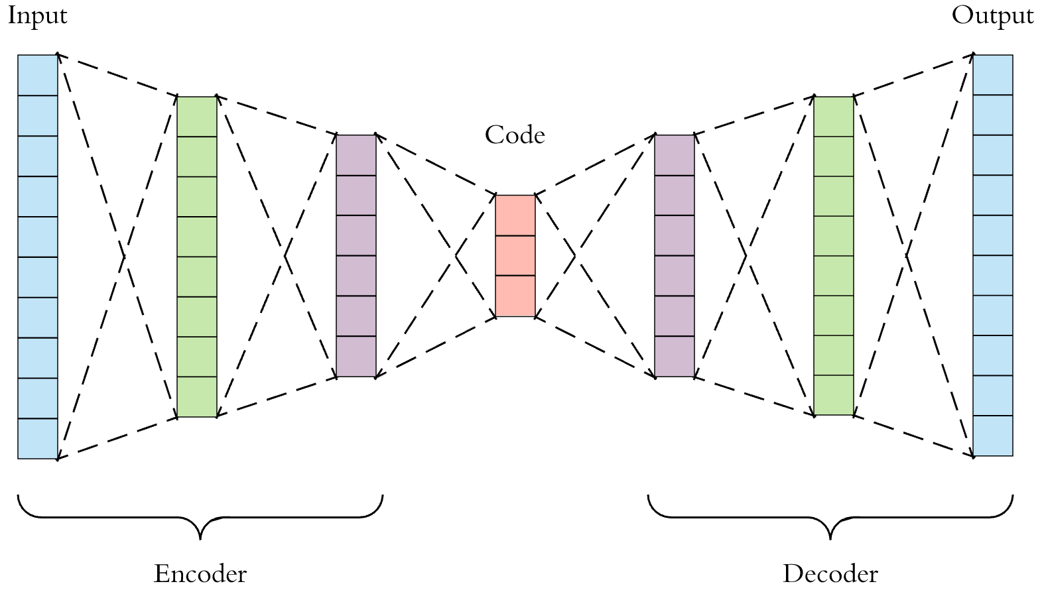}
\caption{Network architecture of Autoencoder}
\end{minipage}
\end{figure}

\subsection*{Background: Word2Vec}
Word2Vec are a type of word embeddings by \cite{DBLP:journals/corr/MikolovSCCD13} that are used to learn dense representations of distributional and semantic context of words in corpus but it's use is not only limited to text. The model borrows from the saying \textit{A word is identified by the company it keeps}, and tries to predict surrounding words of a word (CBOW) or inverse where it tries to predict center word from surrounding words (skip-gram) using a multilayer perceptron.
\begin{figure}[H]
\begin{minipage}{0.48\textwidth}
\centering
\includegraphics[width=\textwidth]{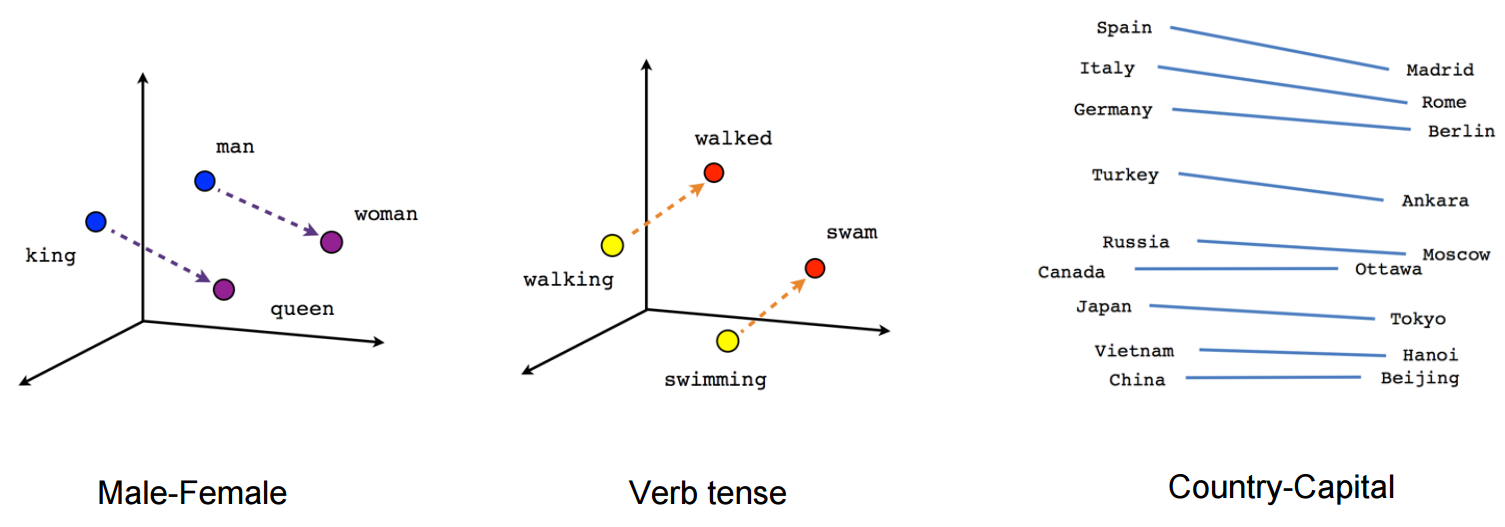}
\caption{Word Vectors latent relations}
\end{minipage}
\end{figure}

\subsection*{Background: Seq2Seq}
Seq2Seq proposed by \cite{sutskever2014sequence} has a similar architecture of autoencoder with the difference that source and target are different. Seq2Seq is used in state of the art neural machine translation systems where the encoder is trained to encode sequential data in source language while decoder is trained to predict target language itself one-time step ahead. 
\begin{figure}[H]
\begin{minipage}{0.48\textwidth}
\centering
\includegraphics[width=\textwidth]{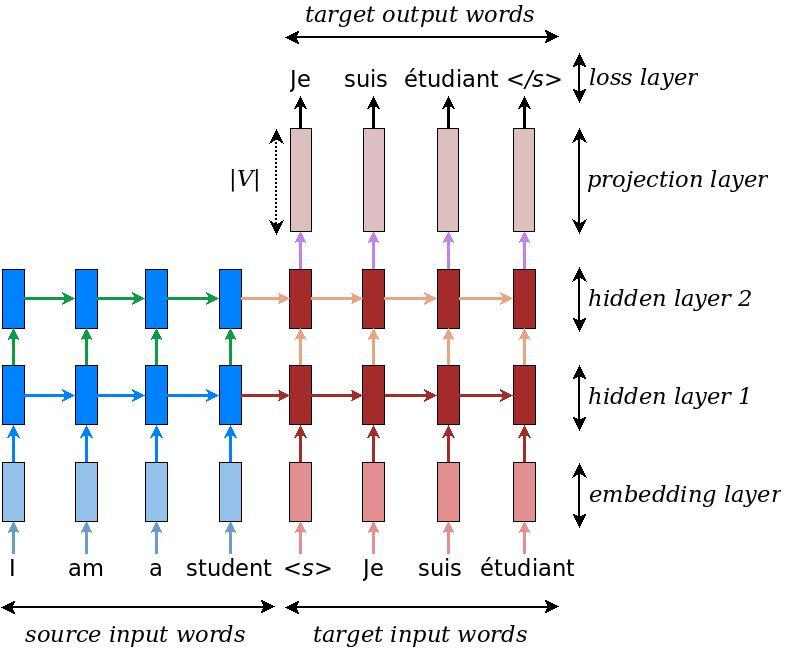}
\caption{Seq2Seq encoder-decoder framework}
\end{minipage}
\end{figure}
The way translation becomes from one language to another becomes possible is that the last state of the encoder is used as the initial state of the decoder which gives it enough context to translate.

\subsection*{Generative Adversarial Networks}
\begin{figure}[H]
\begin{minipage}{0.48\textwidth}
\centering
\includegraphics[width=\textwidth]{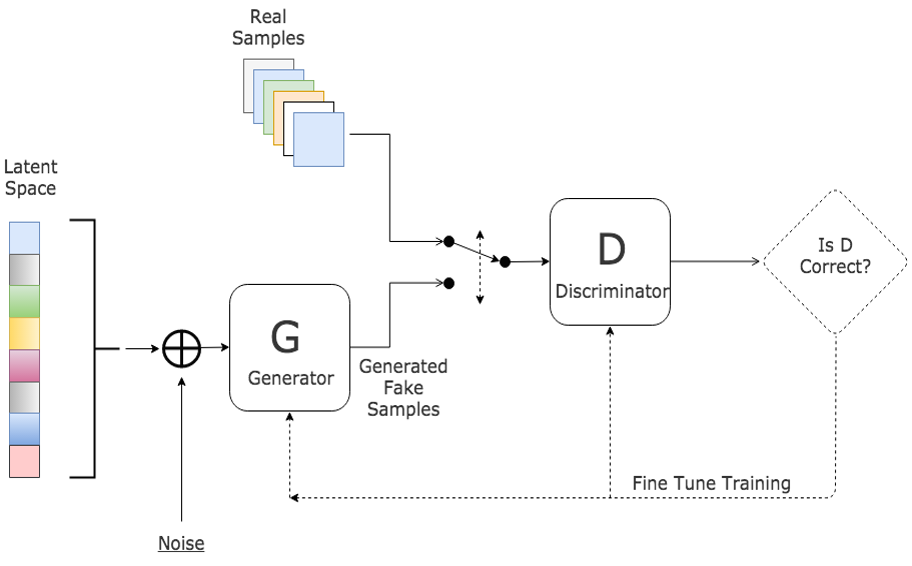}
\caption{GAN: generator-discriminator network}
\end{minipage}
\end{figure}
The main idea behind a GAN \cite{goodfellow2014generative} is to have two competing neural network models. One takes noise as input and generates samples (and so is called the generator). The other model (called the discriminator) receives samples from both the generator and the training data, and has to be able to distinguish between the two sources. These two networks play a continuous game, where the generator is learning to produce more and more realistic samples, and the discriminator is learning to get better and better at distinguishing generated data from real data. These two networks are trained simultaneously, and the hope is that the competition will drive the generated samples to be indistinguishable from real data.

\subsection*{Baseline: Seq2Seq Autoencoder Model}
In the auto-encoder seq2seq model, an encoder is learned to generate intermediate representation of input sequence $X = (x_1, . . . , x_{T_x})$ of length $T_x$. Then a decoder is trained to recover the input $X$ using the intermediate representation. For the style transfer problem, we use the auto-encoder seq2seq model as our base model, since we expect minimum changes from the input to the output. Our intuition is to first learn embeddings of words in a sequence and then encode sequential structure using recurrent neural networks and finally use autoencoders to learn higher representations of our data. Unlike traditional seq2seq we let our source and target be exactly the same based on the intuition that when the model is fully trained and is used for inference, the encoder would encode a different style sentence in the style encoder was trained on and the same with decoder transferring the sentiment effectively. Recall that we are assuming our vocabulary stays the same, we found our intuition to be on the right track from experimental results.

The drawback of this model is that the autoencoder will capture both the style and the content representation which is good if the style we are transferring into also shares the same context e.g. Sentiments where both positive and negative would have the same vocabulary and review context but not generalizable enough for e.g. in case of news to informal style.
\begin{figure*}[ht]
\begin{minipage}{1.0\textwidth}
\centering
\includegraphics[width=\textwidth]{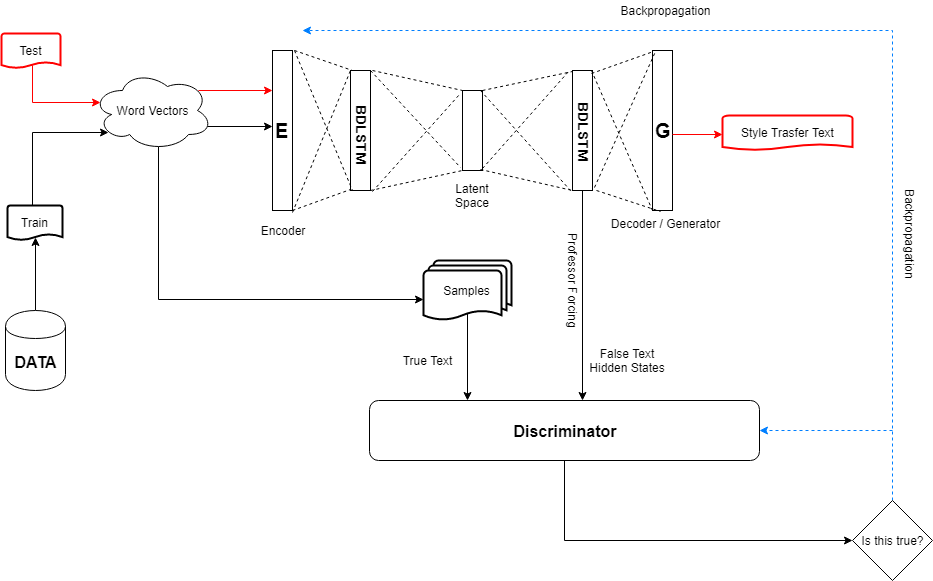}
\caption{Adversarial Seq2Seq Autoencoder Network}
\end{minipage}
\end{figure*}

\subsection*{Reduced Seq2Seq Model}
Another one of our intuitions is that if we let the model see a large vocabulary which it will from the whole dataset, it will learn highly variant style in our case both positive and negative. This would not be ideal for our case since we do not want to build a language model rather a style and content model.

In our experiments, we found that limiting the vocabulary works best for this seq2seq autoencoder, even outperforming full dataset model.

\subsection*{Reversed Seq2Seq Model}
\cite{sutskever2014sequence} found that if we only reverse the source sequence and let target be intact, it reduces the distance between translation words and hence increase memory ability on long sentences while reducing computation time. Reversed model is relevant when using unidirectional LSTMs since bidirectional would automatically consider both directions of a sequence.

In our experiments, we did not find reversing technique to be any superior, and would like to point that it is case specific.

\subsection*{Bidirectional Seq2Seq Model}
Traditional LSTM are unidirectional which means they take only the word appearing after them into context. Adding another layer in parallel that takes the previous word into context and merging/concatenating/averaging both layers allows the model to take both neighbor words into context and is found to be outperforming unidirectional LSTMs.

In our experiments, we found that not only did bidirectional networks plateaued faster but also gave the best accuracy when used in our evaluation model (sentiment classifier) but for our style transfer the bidirectional overtrained itself and was even able to predict exact match unseen data.

\subsection*{Reduced Bidirectional Seq2Seq Model}
After seeing that reduced vocabulary model actually works to our advantage, we tried the same approach of reduced vocabulary with bidirectional LSTMs.

\subsection*{Adversarial Seq2Seq Autoencoder Model}
To separate content from the style we train the seq2seq autoencoder with an adversarial loss where we improve the latent semantic space of the autoencoder to capture a style by training the discriminator to learn to differentiate between a real style using decoder part of autoencoder and a randomly generated fake style.
Since the text decoder outputs discreet values, differentiating on them to back propagate errors would be impossible so we use the last hidden layer of the decoder as input to discriminator and compare that to fake text generated randomly which allows the discriminator to better distinguish original style from the rest.
This network although looks promising on paper was found to be slow to train, given the time constraints we prioritized experimenting more with Seq2Seq Autoencoder.

\section*{Evaluation}
Evaluation plays an important role in style transfer as they provide criteria to compare different models. Automatic evaluation metrics speed up development. In this paper, we will focus on showcasing results of negative to positive transfer for succinctness, in experiments we have found the performance of from positive to negative equal. We divide our evaluation into two types: machine and human evaluation comprising of
\begin{enumerate}
\item Soundness (generated texts being textually entailed with original version)
\item Coherence (free of grammatical errors, proper word usage, etc.)
\item Effectiveness (the generated texts actually match the desired style)

\end{enumerate}

For machine evaluation, BLEU \cite{papineni2002bleu} is a popular evaluation metric in neural machine translation and ROUGE \cite{lin2005recall} is popular in text summarization. But since both of them rely on human-generated text, which in our case is style which can be different for different persons. ROUGE-2 compares percentage bi-gram overlaps between human generated and system generated responses which might be good but in case of sentiment transfer we can simply use state of art classifier to detect whether the transferred sentence successfully transferred the sentiment.
\begin{figure}[H]
\begin{minipage}{0.48\textwidth}
\centering
\includegraphics[width=\textwidth]{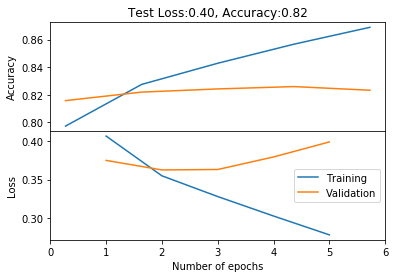}
\caption{Accuracy \& loss of Sentiment Classifier}
\end{minipage}
\end{figure}
Our sentiment classifier is a bidirectional dual layer 1024 units LSTM (dropout 0.2) with a multi-layer perceptron on top trained on 2M equally distributed positive and negative reviews which has an accuracy of 82.6\% on unseen data in just 5 epochs. The train and validation loss intersected in just 2 epochs but we let the network runs for 3 more epochs just to be sure.

\section*{Experiments}
Note that for the purpose of easy evaluation we are framing our problem in terms of sentiment transfer but our training methods are agnostic of the style and are end to end fully unsupervised.

\subsection*{Model}
Before we implement the above model, as a part of understanding of how the basic components work we also implemented the following:
\begin{enumerate}
\item Autoencoders for MNIST dataset and text data where we realized that in case of text we need some network to take the sequential nature of text into account. This is how we developed an intuition of using Seq2Seq Autoencoders. 
\item Word2Vec using noise contrastive estimation to learn word embeddings but quickly realized that in case of non-parallel data we would not have the luxury to pre-train word vectors, so we implement on the fly training of embeddings.
\item Seq2Seq inspired from Google Neural Machine Translation system which includes some of the state of the art techniques like gradient clipping and source sentence reversing.
\item Generative Adversarial Networks on MNIST dataset where we learned how tricky it is to tune GANs as we experience a common phenomenon called \textit{mode collapse} ourselves where generator starts generating only one type of distribution which it thinks is the best way to fool discriminator. To alleviate the problem we used mini-batching to let discriminator see a batch of examples instead of one.
\end{enumerate}

\section*{Results}
\begin{table}[ht]
\label{Models Accuracy Comparison}
\begin{center}
\begin{tabular}{| l | c | c |}
\hline
\textbf{Seq2Seq Models} & \textbf{Loss} & \textbf{Accuracy \%} \\ \hline
Vanilla & 3.12 & 41.70 \\ \hline
Reduced & 2.12 & 66.72 \\ \hline
Reversed & 1.71 & 59.41 \\ \hline
Bidirectional & 2.86  & 45.46 \\ \hline
Reduced Bidirectional & \textbf{1.02}  & \textbf{66.91} \\ \hline
\end{tabular}
\caption{Accuracy of Different Seq2Seq Models}
\end{center}
\end{table}
Evaluating results of all the models using the sentiment classifier we mentioned in the evaluation section, we found \textit{reduced bidirectional} to be the best model for transferring sentiments. Along with that, we have the following observations:
\begin{enumerate}
\item Reducing vocabulary not only increased vocabulary and reduced loss but also trained much much faster as this only used 20k records. Adding bidirectional on top reduced the loss by 100\% which makes reduced bidirectional our best model for sentiment transfer.
\item Reversing the sentences even though did not get us to highest accuracy but if you notice the loss, it not too far behind. This is a completely new finding that we never saw mentioned in literature, we corroborate the findings of \cite{sutskever2014sequence} in the case of Seq2Seq autoencoders as well.
\item Using bidirectional is found record-setting good in general but in case of autoencoders, it is good if the only goal is to encode representation, not really style transfer since it is so good that it overfits the text capturing both content and style in the semantic space which does not serve well for transferring style.
\item Adversarially training the network takes a lot of iterations ($10^5$) and low learning rate ($10^{-4}$), we trained it on a small corpus and got results that verified the architecture is right but to get sensible style transfer meant training on the big dataset which took a lot of time. We think freezing the discriminator layers and training generator on a larger batch would increase learning time. We are keeping that as the future scope.
\item Scope Update: Initially we wanted to transfer in Wikipedia style but given that difference between a sample text and it's Wikipedia counterpart made evaluation tricky, which is why we switched to sentiment transfer on Yelp dataset because the difference in two reviews of 1 vs 5 ratings is stark and can be evaluated easily. Usage of the model for transferring in Wikipedia style should be approached once the sentiment transfer model show promises and the evaluation scheme is robust enough.
\item Coming up with evaluation scheme using a neural classifier fit perfectly in our case when we switched to sentiment transfer.
\end{enumerate}
 
\newpage
\subsection*{Training Speed and Performance}
We noticed a general trend in networks, the higher the number of parameters to train the longer each epoch took. Deepening the networks slowed the training time the most while not really adding much to accuracy score whereas widening the network not only kept the training time from slowly drastically but even added 3 points to accuracy pushing it to 99.83 on train set\%.

Lastly, we want to notify that deep learning is compute intensive, so we kept our hyperparameters in a decent range training everything on an NVIDIA GeForce GTX 1080Ti with tensorflow 1.4, CUDA 8.0, cuDNN 6.1, python 3.6
\begin{enumerate}
\item 2 layers of LSTMs with hidden units to 1024 initialized with a uniform distribution between -0.1 and 0.1 with a seed of 2. Notice in case of bidirectional LSTM the hidden units are 512 which are concatenated as the output of LSTM which makes the number of parameters equal in all models.
\item Fixed number of epochs to 300 to make sure network plateau, we avoided early stopping since gradient descent guarantees local minima not global and gets stuck.
\item Embedding size 300 which is enough to learn a dense representation of our vocabulary size (10k), also embeddings are allowed to be re-trained in each epoch. We are using pretrained embeddings since the original ones were trained on google news which is a formal style of text whereas Yelp reviews are more informal.
\item Batch sizes were calculated so that there are around 512 mini-batches per epoch, this makes sure we have equal mini batches in all reduced and full dataset models.
\item Adam Optimizer which is basically \textit{rmsprop} (suitable for text) with momentum with a Learning Rate: 0.01
\item We are using Greedy Decoder with \textit{GreedyEmbeddingHelper} which uses the \textit{argmax} of the output (treated as logits) and passes the result through an embedding layer to get the next input.
\item We are using a sequence loss which is basically a weighted cross-entropy loss for a sequence of logits.
\item Although LSTMs tend to not suffer from the vanishing gradient problem, they can have exploding gradients, to address that we clipped our gradients at (-5.0, 5.0)
\item Different sentences have different lengths. Most sentences are short (e.g., length 20-30) but some sentences are long (e.g., length \textgreater 100), so a minibatch of 512 randomly chosen training sentences will have many short sentences and few long sentences, and as a result, much of the computation in the minibatch is wasted. To address this problem, we made sure that all sentences within a minibatch were roughly of the same length, which a 2x speedup.
\end{enumerate}

\begin{figure}[H]
\centering
\includegraphics[width=0.5\textwidth]{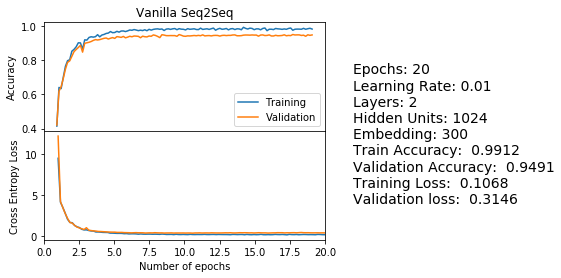}
\includegraphics[width=0.5\textwidth]{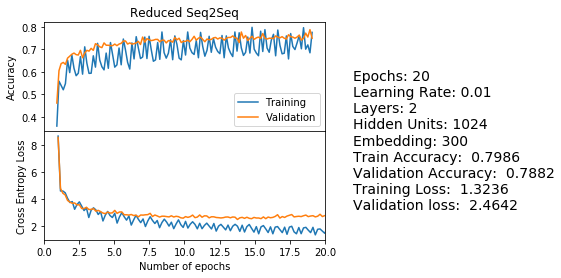}
\includegraphics[width=0.5\textwidth]{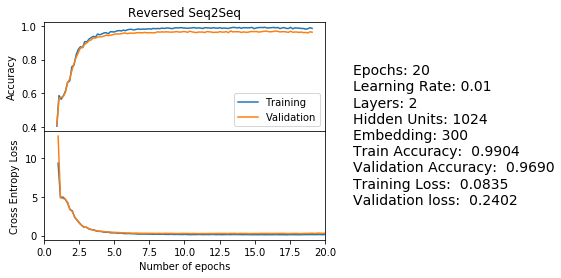}
\includegraphics[width=0.5\textwidth]{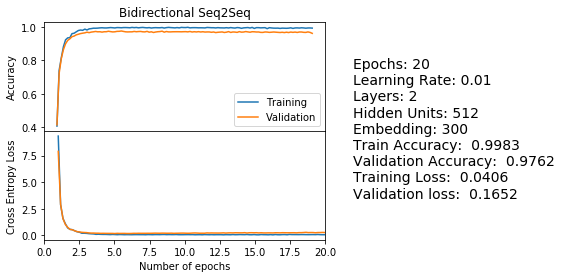}
\includegraphics[width=0.5\textwidth]{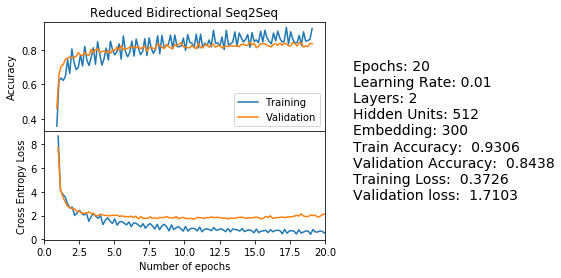}
\caption{Accuracy and Loss measurements for various models}
\end{figure}

\onecolumn
\subsection*{Qualitative Study}
To give people some intuitive sense of how our model performs, we sampled some instances from each style transfer case, and evaluate them three metrics defined in the evaluation section. Below are randomly sampled results of five models in Table 2. We can see that the auto-encoder almost always produces output text that is sound, coherent and effectively transferred style in most of the cases as the input with changing the sentiment from positive to negative or vice versa. Whereas, the other models tend to generate results that replace a few significant words or phrases while changing the sentiment. We were successfully able to do transfer sentiment while preserving most of the common content, thus fulfilling our aim.

\begin{table*}[!h]
\label{Models Accuracy Comparison}
\begin{center}
\begin{tabular}{ l l }
\hline
\textbf{Type} & \textbf{Negative Sentiment Transfer to Positive Sentiment}  \\ 
\hline
Ground Truth & ok never going back to this place again . \\
Vanilla & ok will be going to this place again . \\
Reduced & plus loves going back to this place again . \\
Reversed & this place again place ok find phoenix back . \\
Bidirectional & phenomenal again this location to going going delicious up ! \\
Reduced Bidirectional & this place again this place every time soon \\
\hline
Ground Truth & very disappointed ! \\
Vanilla & very efficient ! \\
Reduced & very impressed ! \\
Reversed & disappointed as ! \\
Bidirectional & beware delightful ! \\
Reduced Bidirectional & roti . \\
\hline
Ground Truth & bad management . \\
Vanilla & pure management . \\
Reduced & truly management . \\
Reversed & honestly night . \\
Bidirectional & good management . \\
Reduced Bidirectional & management ! \\
\hline
Ground Truth & i love the food ... however service here is horrible . \\
Vanilla & i love the food ... this is always and delicious . \\
Reduced & i love the food ... however service here is crazy . \\
Reversed & needless i love the food ... however service is pretty . \\
Bidirectional & they provide is here service however , most the club tacos ! \\
Reduced Bidirectional & i love the food ... however service here is amazing . \\
\hline
Ground Truth & i will never be back . \\
Vanilla & i will definite be back . \\
Reduced & i will be going back . \\
Reversed & back however i will never be \\
Bidirectional & give back be be be again ! \\
Reduced Bidirectional & i will be back . \\
\hline
Ground Truth & the service sucks , management is terrible . \\
Vanilla & the service , folks , is spectacular . \\
Reduced & the service came , management is terrible . \\
Bidirectional & the service great, management better . \\
Reversed & have fees is funny , open service occasions ! \\
Reduced Bidirectional & management is fairly unbelievable , was amazing ! \\
\hline
Ground Truth & they are completely unprofessional and have no experience . \\
Vanilla & they are fully greatly and have made better . \\
Reversed & all here could have and and dig cars ! \\
Reduced & they are completely trendy and have no experience . \\
Bidirectional & have no experience no we are completely trip . \\
Reduced Bidirectional & have myself experience now experience again again area .
\end{tabular}
\caption{Samples from different models}
\end{center}
\end{table*}

\twocolumn
\section*{Conclusion}
We discussed the state of research on the topic of style transfer, the reasons behind lack of research, its applications in all fields of natural language comprehension, and how some of the deep learning techniques like sequence to sequence and autoencoders can be of help. We proposed a model and experimented with different types of architectures, vocabulary size, depth of network as well as the type of cells. We created our own evaluation criterion which is suitable for the task of sentiment classification. Our best model transferred sentiment with 67\% accuracy showing results that even seem reasonable from human perception from the qualitative study. The paper shows a promise on the power of autoencoders combined with seq2seq in multi-domain adaption and calls for future work in using adversarial training or single-encoder for content preservation and multi-decoder to decode in multiple styles at ones.

\section*{Acknowledgement}
We would like to thank Prof. Nik Brown for guiding us and giving us the opportunity to explore this research area. We would also like to thank Prof. Stephen Intille for providing us with access to NVIDIA GPU which helped us reduce training and experimentation time.

\bibliographystyle{apalike}
\bibliography{references}

\end{document}